\documentclass[letterpaper, 10 pt, conference]{ieeeconf}

\IEEEoverridecommandlockouts                              

\usepackage{graphics} 
\usepackage{epsfig} 
\usepackage{amsmath} 
\usepackage{amssymb}  
\usepackage{algpseudocode}
\usepackage{algorithm}
\usepackage{url}

\usepackage{caption}
\usepackage{subcaption}
\usepackage{color}

\let\labelindent\relax 
\usepackage{enumitem}

\newtheorem{theorem}{Theorem}

\newtheorem{definition}{Definition}    
    
\title{\LARGE \bf
Actor-Critic Physics-Informed Neural Lyapunov Control
}

\author{Jiarui Wang$^{1}$ and Mahyar Fazlyab$^{2}$
\thanks{$^{1}$Department of Computer Science,
        Johns Hopkins University, 3400 N Charles St, Baltimore, MD 21218
        {\tt\small jwang486@jhu.edu}}%
\thanks{$^{2}$Department of Electrical and Computer Engineering,
        Johns Hopkins University, 3400 N Charles St, Baltimore, MD 21218
        {\tt\small mahyarfazlyab@jhu.edu}}%
}

\begin{document}

\maketitle

\begin{abstract}
Designing control policies for stabilization tasks with provable guarantees is a long-standing problem in nonlinear control. A crucial performance metric is the size of the resulting region of attraction, which essentially serves as a robustness ``margin'' of the closed-loop system against uncertainties. In this paper, we propose a new method to train a stabilizing neural network controller along with its corresponding Lyapunov certificate, aiming to maximize the resulting region of attraction while respecting the actuation constraints. Crucial to our approach is the use of Zubov's Partial Differential Equation (PDE), which precisely characterizes the true region of attraction of a given control policy. Our framework follows an actor-critic pattern where we alternate between improving the control policy (actor) and learning a Zubov function (critic). Finally, we compute the largest certifiable region of attraction by invoking an SMT solver after the training procedure. Our numerical experiments on several design problems show consistent and significant improvements in the size of the resulting region of attraction. 
\end{abstract}


\section{Introduction}
\label{sec:introduction}
Neural network control policies have shown great promise in model-based control and outperformed existing design methods, owing to their capability to capture intricate nonlinear policies \cite{li2020aggressive} \cite{yang2013neural} \cite{zhang2022neural} \cite{kaufmann2023champion}. For example, neural network control policies trained with reinforcement learning algorithms have been able to outperform human champions in drone racing \cite{kaufmann2023champion}.
Despite the outstanding empirical performance, the application of neural network policies in physical systems is of concern due to a lack of stability and safety guarantees. 

Certificate functions such as Lyapunov functions provide a general framework for designing nonlinear control policies with provable guarantees. For instance, a quadratic Lyapunov function can be derived using a Linear Quadratic Regulator (LQR) \cite{kwakernaak1972linear} for linear systems, while a polynomial Lyapunov function can be obtained through Sum-of-Squares (SoS) optimization \cite{ahmadi2016some} for polynomial systems. However, as systems become more complex and nonlinear, the automated construction of Lyapunov functions becomes an increasingly crucial research focus within control theory  \cite{Verhoek_2023,abate2020formal,ahmed2020automated,dawson2023safe}.

Learning-based methods combined with neural networks have been a promising alternative to optimization-based methods in constructing Lyapunov certificates for autonomous systems 
\cite{abate2020formal}\cite{gaby2022lyapunov}\cite{grune2020computing}
or co-learning a neural network control policy together with a neural Lyapunov function \cite{chang2019neural,jin2020neural,zhou2022neural}. In these learning-based methods, a typical approach is to minimize the violation of the Lyapunov conditions at a finite number of sampled states, and then invoke a verifier such as SMT solvers \cite{barrett2018satisfiability} to extend the validity of the learned certificate beyond the sampled states. Although applicable to more general nonlinear systems, the choice of the training loss function is critical to the learner's success. For example, for the case of stability analysis, we are interested in certifying the largest region of attraction, which is missing in a training objective that merely penalizes the Lyapunov condition violations. 

\subsubsection*{Our Contribution} In this paper, we propose an algorithmic framework for co-learning a neural network control policy and a neural network Lyapunov function for actuation-constrained nonlinear systems. Our starting point is to use a physics-informed loss function based on Zubov's Partial Differential Equation (PDE), the solution of which characterizes the ground domain of attraction (DoA) for a given control policy. 
Our framework then follows an actor-critic pattern where we alternate between learning a Zubov function, by minimizing the PDE residual, 
and improving the control policy, by minimizing its Lie derivative akin to Sontag's formula \cite{sontag2013mathematical}. 
To improve the efficiency of the method, we provide an implementation that essentially runs the actor and critic updates in parallel. Finally, we propose a novel method based on the softmax function to enforce polytopic actuation constraints without any projection layer. 
Our numerical experiments on several stabilization tasks show that our approach can enlarge the DoA significantly compared to state-of-the-art techniques.
The source code is available at \url{https://github.com/bstars/ZubovControl}.

\subsection{Related Work}

\subsubsection*{Learning Lyapunov Functions} There is a large body of work on learning Lyapunov functions for autonomous systems \cite{abate2020formal} \cite{gaby2022lyapunov}
\cite{grune2020computing} \cite{boffi2021learning}.
In the work of Abate et al \cite{abate2020formal}, they proposed a counter-example guided framework to learn a neural Lyapunov function. 
A common feature of these methods is that the level set of the learned Lyapunov function does not match the true DoA, leading to a conservative estimation of the DoA despite using neural networks. 
Liu et al \cite{raissi2019physics} used Zubov's PDE as a training loss to capture the maximum region of attraction for autonomous systems. Our method also utilizes Zubov's equation to \emph{co-learn} the control policy and the stability certificate together. 

\subsubsection*{Co-learning Lyapunov Functions and Policies} 
Chang et al. \cite{chang2019neural} proposed a counter-example guided method similar to \cite{abate2020formal} to co-learn a Lyapunov function and a control policy that jointly satisfy the Lyapunov conditions for nonlinear systems.  In their work, they add a regularizer to the loss function to enlarge the DoA. The method in \cite{chang2019neural} has also been extended to safety certificates \cite{jin2020neural}, unknown nonlinear systems \cite{zhou2022neural} and discrete-time systems \cite{wu2024neural}. In our method, regularization is unnecessary as the Zubov equation captures the true DoA.

\subsubsection*{Actor-critic and Policy Iteration}
Our framework is also inspired by the actor-critic method \cite{konda1999actor} \cite{sutton2018reinforcement} in reinforcement learning. Actor-critic methods consist of a control policy and a critic that evaluates the value function of the policy. Then the algorithm switches between evaluating the current policy and improving the policy by maximizing the value function. 

Actor-critic methods can be seen as an approximation of policy iteration \cite{sutton2018reinforcement}. There have also been works that combine neural Lyapunov functions with policy iteration to learn formally verifiable control policies. In the work of \cite{meng2024physics}, they use policy iteration to learn control policies for control-affine systems. The policy evaluation part is done by representing the value function by a physics-informed neural network and minimizing the violation of the Generalized Hamilton-Jacobi-Bellman equation \cite{leake1967construction}, and the policy improvement is done by minimizing the value function. 
Our method also follows a policy-evaluation-policy-improvement paradigm as in the actor-critic method and policy iteration.

The rest of the paper is organized as follows. In$\S$\ref{section:Background and Problem Statement}, we provide background and problem statement.  In $\S$\ref{section:Physics-Informed Actor-Critic Learning}, we describe how to co-learn a control policy and a Lyapunov function with a physics-informed neural network and the verification procedure after training. We provide numerical experiments in $\S$\ref{section: Numerical Experiment}.

\subsection{Notation}
We denote $n$-dimensional real vectors as $\mathbb{R}^n$. For $x \in \mathbb{R}^n$, $\|x\|$ denotes the Euclidean norm. $[x_1,...,x_n]$ represents a matrix with columns $x_1, ...,x_n$. For a set $\mathcal{A}$, $\partial \mathcal{A}$ denotes the boundary of $\mathcal{A}$ and $\mathcal{A} \setminus \{0\}$ denotes the set $\mathcal{A}$ excluding the single point $0$. 
For a vector $x \in \mathbb{R}^n$, we use the notation $(x)_+=\max(x,0)$. $\mathbf{sg}(\cdot)$ stands for ``stop gradient'', meaning that the argument in $\mathbf{sg}(\cdot)$ operator is treated as constant. (e.g.,  $\frac{d}{d x} (x \cdot \mathbf{sg}(ax)  ) = ax$.

\section{Background and Problem Statement}  \label{section:Background and Problem Statement}

\subsection{Lyapunov Certificates}
Consider the autonomous nonlinear continuous-time dynamical system
\begin{align} \label{eq:autonomous}
    \dot{x} = f(x) 
\end{align}
where $x \in \mathcal{D} \subseteq \mathbb{R}^n$ is the state and $f \colon \mathcal{D} \to \mathbb{R}^n$ is a Lipschitz continuous function. 
For an initial condition $x_0 \in \mathcal{D}$, we denote the unique solution by $x(t; x_0)$, which we assume exists for all $t \geq 0$. We assume that the origin is an equilibrium of the system,  $f(0)=0$. 

\begin{definition}[Stability\cite{Haddad_Chellaboina_2008}]
    The zero solution $x(t)=0$ to (1) is stable if for any $\epsilon > 0$, there exists $\delta>0$ such that if $\|x_0\|<\delta$, then $\|x(t; x_0)\| < \epsilon$ for all $t>0$.
\end{definition}

\begin{definition}[Asymptotic Stability\cite{Haddad_Chellaboina_2008}]
The zero solution $x(t)=0$ to (1) is locally asymptotically stable if it is stable and there exists a $\delta > 0$ such that if $\|x_0\| < \delta$, then $\lim_{t \to \infty} \|x(t; x_0)\|=0$.
\end{definition}

\begin{definition}[Domain of Attraction~\cite{Haddad_Chellaboina_2008}] If the zero solution $x(t)=0$ to (1) is asymptotically stable, then the domain of attraction (DoA) is given by 
$
\mathcal{A} := \big\{ x_0 \in \mathcal{D} : \lim_{t \to \infty} \| x(t; x_0)\| = 0 \big\}
$. 
\end{definition}

A common way to certify the stability of an equilibrium and estimate the DoA is via Lyapunov functions.

\begin{theorem}[Lyapunov Stability\cite{Haddad_Chellaboina_2008}]
Consider the dynamical system in (1). If there exists a continuously differentiable function $V \colon \mathcal{D} \rightarrow \mathbb{R}$ satisfying the following conditions 
\begin{subequations} \label{eq: lyapunov conditions}
    \begin{align}
        &V(0) = 0 \\
        &V(x)  > 0 ~ \forall ~ x \in \mathcal{D} \setminus \{0\} \\
        &\nabla V(x)^\top f(x) < 0 ~ \forall x \in \mathcal{D} \setminus \{0\}
    \end{align}
\end{subequations}
then the zero solution $x(t)=0$ is asymptotically stable.
\end{theorem}

Any sub-level set $\mathcal{D}_\beta = \{ x : V(x) \leq \beta \} \subseteq \mathcal{D}$ containing the origin is a subset of DoA \cite{Haddad_Chellaboina_2008}, and the largest $\mathcal{D}_\beta$ can be calculated by maximizing $\beta$ subject to $\mathcal{D}_\beta \subseteq \mathcal{D}$.

\subsection{Problem Statement}

Consider the nonlinear continuous-time dynamical system
\begin{align}
    \dot{x} = f(x, u) 
\end{align}
where $x \in \mathcal{D} \subseteq \mathbb{R}^n$ is the state, $u \in \mathcal{U}$ is the control input, $\mathcal{U} \subseteq \mathbb{R}^m$ is the actuation constraint set, and $f \colon \mathcal{D} \times \mathcal{U} \to \mathbb{R}^n$ is a Lipschitz continuous function. This paper assumes that $\mathcal{U}$ is a convex polyhedron.

Given a locally-Lipschitz control policy $\pi(\cdot)$ and an initial condition $x_0 \in \mathbb{R}^n$, we denote the solution of the closed-loop system 
$\dot{x} = f(x, \pi(x)) = f_{cl}(x)$
at time $t \geq 0$ by $x(t, x_0; \pi)$. In this paper, we are interested in learning a nonlinear control policy $\pi_{\gamma}(\cdot)$, parameterized by a neural network with trainable parameters $\gamma$, with the following desiderata:
\begin{itemize}
    \item[1.] The control policy respects the actuation constraints, $\pi_{\gamma}(x) \in \mathcal{U} ~ \forall ~ x \in \mathcal{D}$; 
    \item[2.] The zero solution of the closed-loop system $\dot{x} = f(x, \pi_\gamma(x))$ is asymptotically stable; and
    \item[3.] The DoA of the zero solution is maximized. 
\end{itemize}

\subsection{Neural Lyapunov Control}
A typical learning-enabled approach to designing stabilizing controllers is to parameterize the Lyapunov function candidate $V_\theta$ and the controller $\pi_{\gamma}$ with neural networks and aim to minimize the expected violation of Lyapunov conditions \eqref{eq: lyapunov conditions}
\begin{align}\label{eq: Lyapunov risk}
    \min_{\theta, \gamma} \! E_{x \sim \rho} \! \big[ 
        V_\theta(0)^2 \!+\! 
        (-V_\theta(x))_{+} \!+\! \big(\nabla_x V_\theta(x)^\top f(x, \pi_\gamma(x))\big)_{+},
    \big]
\end{align}
where $\rho$ is a predefined sampling distribution with support on $\mathcal{D}$ \cite{chang2019neural}. There are potentially infinitely many Lyapunov function candidates that minimize the expected loss, including the shortcut solution $V_{\theta}(x)=0 ~\forall x \in  \mathcal{D}$. To avoid this shortcut and promote large DoA, we must use regularization, e.g., $( \|x\|_2 - \alpha V_\theta(x) )^2$ \cite{chang2019neural}. However, poorly chosen regularizers can lead to a mismatch between the shape of level sets of the learned Lyapunov function and the true DoA. Motivated by this drawback, we will incorporate a physics-informed loss function to directly capture the true DoA. 

\section{Physics-Informed Actor-Critic Learning} \label{section:Physics-Informed Actor-Critic Learning}
\subsection{Maximal Lyapunov Functions and Zubov's Method}
To characterize the DoA exactly, we can use a variation of the Lyapunov function, the \textit{maximal Lyapunov function} \cite{vannelli1985maximal}. 

\begin{theorem}[Maximal Lyapunov Function\cite{vannelli1985maximal}] \label{theorem:maximal}
Suppose there exists a set $\mathcal{A} \subseteq \mathcal{D}$ containing the origin in its interior, a continuously differentiable function $V \colon \mathcal{A} \rightarrow \mathbb{R}$, and a positive definite function $\Phi$ satisfying the following conditions
\begin{subequations}
\begin{align}
    &V(0)=0,  V(x)>0 ~ \forall x \in \mathcal{A} \setminus \{0\} \\
    &\nabla V(x)^\top f(x)=-\Phi(x) ~ \forall x \in \mathcal{A} \\
    &V(x) \to \infty \text{ as } x \to \partial \mathcal{A} \text{ or } \|x\| \to \infty
\end{align}
\end{subequations}
Then $\mathcal{A}$ is the DoA for the system in (\ref{eq:autonomous}).
\end{theorem}

One can verify that the function
\begin{align} \label{eq:integral}
    V(x_0) = 
    \begin{cases}
        \int_0^{\infty} \|x(t; x_0)\| dt & \text{if the integral converges} \\
        \infty & \text{otherwise}
    \end{cases}
\end{align}
satisfies the conditions of Theorem 2 with $\Phi(x)=\|x\|$  \cite{vannelli1985maximal}.

Due to the presence of infinity in the integral, the above maximal Lyapunov function is hard to represent by function approximators such as neural networks. Another way to build a Lyapunov function and characterize the DoA is using Zubov's Theorem \cite{Haddad_Chellaboina_2008} \cite{zubov1961methods}.

\begin{theorem}[Zubov's Theorem~\cite{zubov1961methods}] \label{theorem:zubov}
Consider the nonlinear dynamical system in (\ref{eq:autonomous}). Let $\mathcal{A} \subseteq \mathcal{D}$ and assume there exists a continuously differentiable function $W \colon \mathcal{A} \rightarrow \mathbb{R}$ and a positive definite function $\Psi : \mathbb{R}^n \rightarrow \mathbb{R}$ satisfying the following conditions,
\begin{subequations}
\begin{align}
    &W(0) =0 \label{eq: Zubov zero at zero}\\
    & 0 < W(x) < 1 ~ \forall x \in \mathcal{A} \setminus \{ 0\} \\
    & W(x) \to 1 \text{ as } x \to \partial \mathcal{A} \text{ or } \|x\| \to \infty \label{eq: Zubov BC}\\
    & \nabla W(x)^\top f(x) = - \Psi(x)(1 - W(x)). \label{eq: Zubov}
\end{align}
\end{subequations}
\end{theorem}
Then $x(t)=0$ is asymptotically stable with DoA $\mathcal{A}$.

\begin{figure}
\centering
\includegraphics[width=0.33\textwidth]{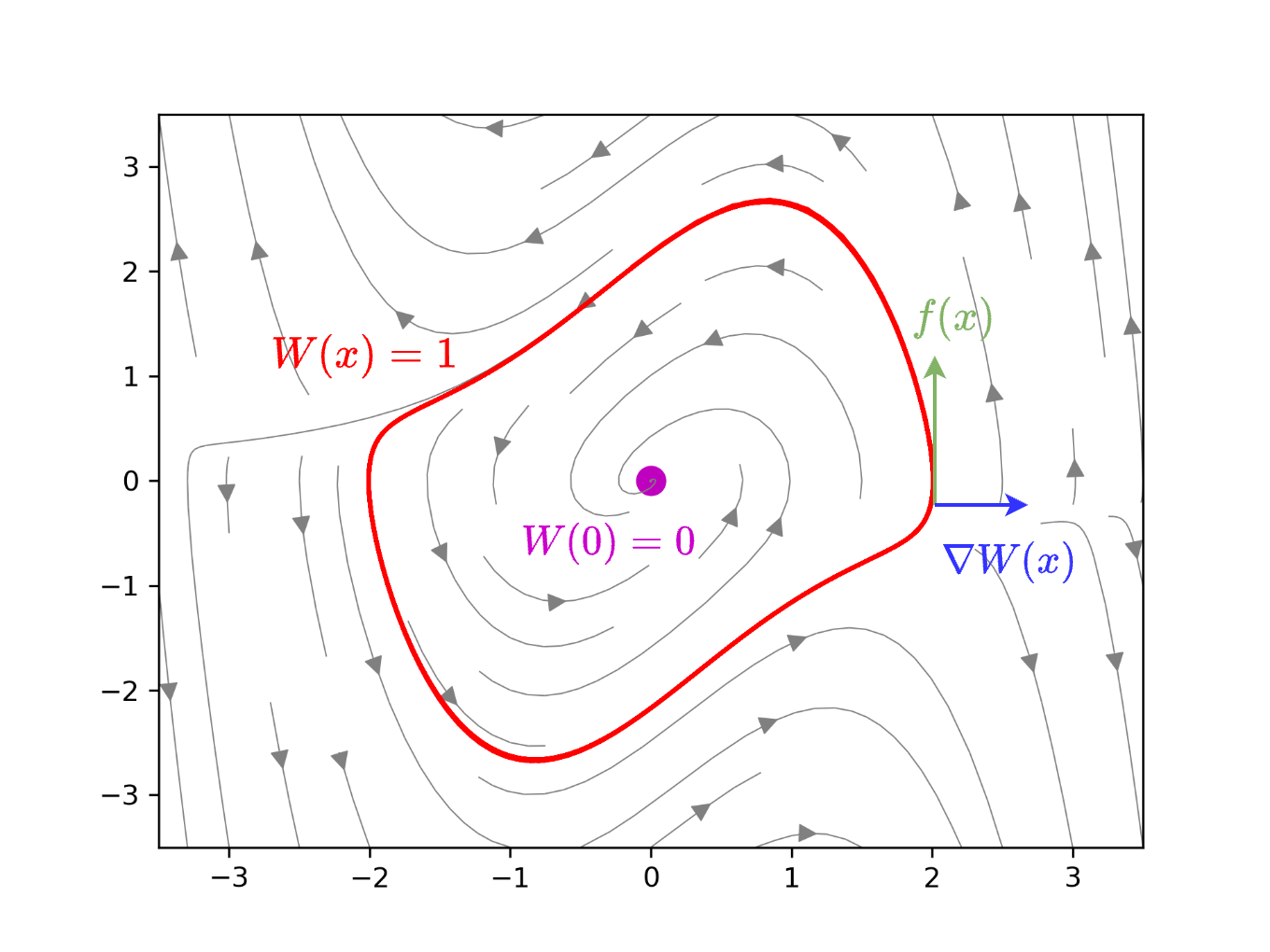}
\caption{The level set $\{x : W(x)=1\}$ obtained by the Zubov PDE characterizes the boundary of the true DoA.}
\label{fig:zubov}
\end{figure}

Intuitively, the Zubov PDE \eqref{eq: Zubov} ensures that the Lie derivative of $W$ on the boundary of $\mathcal{A}$ is zero, a property which precisely renders $\mathcal{A}$ the domain of attraction--see Figure \ref{fig:zubov} for an illustration. 

The maximal Lyapunov function $V(\cdot)$ of Theorem \ref{theorem:maximal} and the Zubov function $W(\cdot)$ of Theorem \ref{theorem:zubov} can be related by \cite{kang2023data}
\begin{subequations}
\begin{align}
    W(x) &= \mathrm{tanh}(\alpha V(x)) \label{eq:tanh_relation}\\
    V(x) &= \frac{1}{2\alpha} \log\big( \frac{1 + W(x)}{1-W(x)}\big) \\
    \Psi(x) &= \alpha (1 + W(x))\Phi(x)
\end{align}
\end{subequations}
where $\alpha>0$. In the rest of the paper, we choose $\Phi(x) = \|x\|_2$.

\subsection{Learning the Zubov Function (Critic)} \label{section:Learning Zubov Function}
Suppose we have a \emph{fixed} policy $\pi_{\gamma}(\cdot)$ and the corresponding closed-loop system $\dot{x}=f(x,\pi_{\gamma}(x))$. Let $V(\cdot)$ be chosen as in \eqref{eq:integral}. The relationship \eqref{eq:tanh_relation}, combined with (\ref{eq: Zubov}), enables us to learn the Zubov function $W_{\theta}(x) \approx \mathrm{tanh}(\alpha V(x))$, which can be interpreted as an evaluation of the policy $\pi_\gamma$. Here $W_{\theta}(\cdot)$ is a neural network with $\mathrm{sigmoid}$ activation functions at the output layer \cite{liu2023physics} such that $W_{\theta}(x) \in (0,1) \ \forall x \in \mathcal{D}$. 
To train $W_{\theta}(\cdot)$, we use the following loss function,
\begin{align}
    L(\theta) = L_z(\theta) + L_r(\theta) + L_p(\theta) 
\end{align}
where
\begin{subequations}
\begin{align}
    L_z(\theta) &= W_{\theta}(0)^2 \\
    L_r(\theta) 
    &= \frac{1}{M} \sum_{i=1}^M
    \big( W_{\theta}(x_i) - \mathrm{tanh}(\alpha V(x_i) \big)^2 \label{eq:V_loss}\\
    L_p(\theta) &= \frac{1}{K} \sum_{i=1}^K 
    \bigg[
        \nabla_x W_{\theta}(x_i)^\top f(x_i, \pi_{\gamma}(x_i)) + \\
        &\qquad \qquad \alpha (1-W_{\theta}(x_i))(1+W_{\theta}(x_i)) \Phi(x_i)
    \bigg]^2 \notag
\end{align}
\end{subequations}The first and second terms, $L_z(\theta)$ and $L_r(\theta)$, penalize the violation of \eqref{eq: Zubov zero at zero} and \eqref{eq:tanh_relation}, respectively.  We note that to evaluate $V(\cdot)$ that appears in this loss, we simulate the closed-loop system from $M$ initial conditions $\{x_i\}_{i=1}^{M}$. 
Finally, the third term $L_p(\theta)$ is the physic-informed part to minimize the residual of the Zubov PDE in \eqref{eq: Zubov} at $K$ sampled states.

\subsection{Policy Improvement (Actor)} \label{section:Policy Improvement}
Given a policy $\pi_\gamma(\cdot)$, the corresponding gound-truth Zubov function is $W(x) = \tanh(\alpha V^{\pi_\gamma}(x))$ where $V^{\pi_\gamma}(x)$ is defined as in \eqref{eq:integral} for the autonomous system $\dot{x} = f(x, \pi_{\gamma}(x))$. We can improve the policy to drive the state to $x=0$ by minimizing the Lie derivative of $W$ along $f$ akin to Sontag's formula \cite{sontag2013mathematical}. However, since we cannot differentiate $W(x)$ with respect to $x$, we use $W_{\theta}(\cdot)$ as a proxy of the true $W$, which results in the following loss function 
\begin{align*}
    L_c(\gamma) = \frac{1}{K} \sum_{i=1}^K \bigg[ \frac{\nabla_x W_{\theta}(x_i)^\top}{\|\nabla_x W_{\theta}(x_i)\|} f(x_i, \pi_{\gamma}(x_i)) \bigg]
\end{align*}
Here we normalize the gradient $\nabla_x W_\theta(x)$ so that the training samples are equally weighted in both the steep regions and flat regions of $W_\theta(\cdot)$. 

\subsection{Actor-Critic Learning}
We now combine the critic of $\S$\ref{section:Learning Zubov Function} and the actor of $\S$\ref{section:Policy Improvement} to co-learn the controller and the Zubov function. We first define $R_1 \subseteq \mathcal{D}$ to be the region from which we sample $\{x_i\}$ and define the region $R_2 = \{ ax : x \in R_1\}$ ($a>1$) as the region of interest. In this paper we choose $a=2$.
Since the controller is randomly initialized, the sampled trajectories might diverge. To prevent this, we add a loss function as follows,
\begin{align*}
    L_b(\theta) = \frac{1}{K} \sum_{x'_i \in \partial R_2, i=1,...,K} \big| W_{\theta}(x'_i) - 1 \big|
\end{align*}
Since $W(\cdot)$ is constrained to be in $[0,1]$ and the actor is trained to minimize the Lie derivative of ${W}$, this loss prevents the states from going to $\partial R_2$, where $W(\cdot)$ takes the maximum value $1$.
This loss function stabilizes the training at the beginning. 

Combining policy evaluation and policy improvement, we can co-learn a controller and a Zubov function in an actor-critic fashion by minimizing the following loss function
\begin{align} \label{eq:total loss}
    L(\theta, \gamma) &= \lambda_0 W_{\theta}(0)^2 \\ \notag
    &~~ + \frac{1}{M} \sum_{x_i \in R_1, i=1,..,M} 
    \big( W_{\theta}(x_i) - \mathrm{tanh}(\alpha V(x_i) \big)^2 \\ \notag
    &~~ + \frac{1}{K} \sum_{x_i \in R_1, i=1,..,K} 
    \bigg[
        \nabla_x W_{\theta}(x_i)^\top f(x_i, \mathbf{sg}(\pi_{\gamma}(x_i)) \\  \notag
        &~~~~~~~~~~~~~~~~~~ +
        \alpha (1-W_{\theta}(x_i))(1+W_{\theta}(x_i)) \Phi(x)
    \bigg]^2 \\ \notag
    &~~ + \frac{\lambda_c}{K} \sum_{x_i \in R_1, i=1,..,K}  \bigg[ \mathbf{sg}(
         \frac{\nabla_xW_{\theta}(x_i)^\top}{\|\nabla_xW_{\theta}(x_i)\|}
    ) f(x_i, \pi_{\gamma}(x_i)) \bigg]  \\ \notag
    &~~ + \frac{\lambda_b}{K} \sum_{x'_j \in \partial R_2, j=1,...,K} \big| W_{\theta}(x'_j) - 1 \big|
\end{align}
Note that rather than alternating between updating $W_\theta(\cdot)$ and $\pi_\gamma(\cdot)$ with the other one fixed, we use the $\mathbf{sg(\cdot)}$ operator to enable a simultaneous update of $W_\theta(\cdot)$ and $\pi_\gamma(\cdot)$. We outline the overall method in Algorithm \ref{alg:cap}.

\begin{algorithm}[H]
\caption{Physics-Informed Neural Lyapunov Control}\label{alg:cap}
\begin{algorithmic}
\State Randomly initialize $W_{\theta}$ and $\pi_{\gamma}$
\For{$n=1,...,T$}
    \State Randomly sample $x_1, ..., x_K$ from $R_1$
    \State Randomly sample $x'_1, ... x'_K$ from $\partial R_2$
    \State Simulate trajectories $x(t, x_i; \pi_\gamma)$ (with RK-4 integrator)
    \State Estimate $V(x_i) \approx \int_t \|x(t, x_i; \pi)\| dt$ for $i=1,..,M$
    \State Take a gradient step to minimize $L(\theta, \gamma)$ in (\ref{eq:total loss})
\EndFor
\end{algorithmic}
\end{algorithm}

\subsection{Actuation Constraints}
To respect actuation constraints, one approach is to append a Euclidean projection layer to a generic neural network $u_\gamma(\cdot)$, resulting in the control policy
\begin{align} \label{eq: euclidean project actuation}
    \pi_{\gamma}(x)=\mathrm{Proj}_{\mathcal{U}} [u_{\gamma}(x)] = \mathrm{argmin}_{u' \in \mathcal{U}} \|u'-u_{\gamma}(x)\|_2^2
\end{align}
For convex $\mathcal{U}$, we can compute the derivative $D_{\gamma} u_{\gamma}(x)$, which is needed to learn $\gamma$, using differentiable convex optimization layers \cite{agrawal2019differentiable}. This approach is particularly efficient for box constraint sets, $\mathcal{U} = \{u : \underline{u} \leq u \leq \bar{u}\}$, for which the projection can be computed in closed form. However, for general convex polyhedrons, the projection has to be solved numerically. To avoid this, our innovative solution is to span the actuation constraint set $\mathcal{U}$ using a convex combination of its vertices. Formally, suppose $V = [v_1 \ \cdots v_M]$ is the matrix of vertices of $\mathcal{U}$. We then parameterize the control policy as
\begin{align}
    \pi_{\gamma}(x) \!=\! V \cdot \mathrm{softmax}(u_{\gamma}(x)) \!=\! \sum_{i=1}^{M} \! V_i \mathrm{softmax}(u_{\gamma}(x))_i
\end{align}
where the vector-valued softmax function simply generates the coefficients of the convex combination of columns of $V$. In contrast to \eqref{eq: euclidean project actuation}, this parameterization eliminates the need to compute any numerical optimization sub-routine. 

\subsection{Verification}
To formally verify the learned certificate, let $c \in (0,1)$. Our goal is to verify the following three conditions
\begin{align} \label{eq:SMT ours}
\begin{cases}
    W_\theta(x)>W_\theta(0) ~ \forall x \in \{x \in R_2 \mid W_\theta(x)<c, \ \|x\| \geq \epsilon\} \\
    \dot{W}_\theta(x)<0 ~ \forall x \in \{x \in R_2 \mid W_\theta(x)<c, \ \|x\| \geq \epsilon\} \\
    W_\theta(x)>c ~\forall x \in \partial R_2
\end{cases}
\end{align}
Since $W_{\theta}$ can have disjoint sub-level sets, we only consider sub-level sets contained in $R_2$.

The first and second conditions verify the Lyapunov conditions on the set $\mathcal{D} = R_2 \cap \{ x: W_{\theta}(x) < c \}$. We ignore a smaller neighbor around the origin to avoid numerical errors. However, the first two conditions do not imply that $\mathcal{D}$ is the domain of attraction. We also need to verify the third condition, which implies that the set $\mathcal{D}$ is strictly contained in $R_2$ so that a trajectory does not leave $R_2$ with negative $\dot{W}_\theta$ along the trajectory. If these conditions are satisfied, then the function $W_\theta(x) - W_\theta(0)$ is a Lyapunov function in the region $R_2 \cap \{ x: W_{\theta}(x) < c \}$.

\section{Numerical Experiments} \label{section: Numerical Experiment}
We demonstrate the effectiveness of our method on various nonlinear control problems. We first run our method on Double Integrator and Van der Pol and compare the verified DoA obtained by our method with that of the LQR controller. We then test our method on Inverted Pendulum and Bicycle Tracking and compare it to both LQR and Neural Lyapunov Control \cite{chang2019neural}\footnote{The code repository of \cite{chang2019neural} does not support the first two examples.}. Across all examples, we use the same hyperparameter $M=8, K=64, \lambda_0=5, \lambda_c = 0.5$, and $\lambda_b = 5$ for the loss function. Other hyperparameters and verified sublevel set can be found in Table \ref{table:hyperparameters}. The effect of different choices of hyperparameters can be found in the preprint version \cite{wang2024actorcritic}. In all experiments, we assumed that the actuation constraint is $u \in \mathcal{U} = \{ u | ~\|u\|_{\infty} \leq 1 \}$ and the learned controller satisfies this constraint by using $\mathrm{tanh}$ as the activation function in the output layer.

To find the DoA of LQR under actuation constraints, we first solve the Riccati equation of the linearized system with $Q=I, R=I$ to obtain the solution $P$ and control matrix $K$, and then we verify the following conditions,
\begin{align} \label{eq:SMT LQR}
\begin{cases}
    x^\top P\dot{x} < 0 ~ \forall x \text{ s.t } x^\top Px < c_{lqr} \text{ and } \|x\| \geq \epsilon\\ \|Kx\|_{\infty} \leq 1 ~ \forall x \text{ s.t } x^\top Px < c_{lqr} \text{ and } \|x\| \geq \epsilon,
\end{cases}
\end{align}
where $\dot{x}$ is given by the nonlinear closed-loop dynamics. 

We used the SMT solver dReal \cite{gao2013dreal} to verify the conditions (\ref{eq:SMT ours}) for the neural network controller and (\ref{eq:SMT LQR}) for the LQR controller. To find the largest DoA for both LQR and our method, we perform bisection to find the largest $c_{lqr}$ ($c$) such that conditions \eqref{eq:SMT ours}, \eqref{eq:SMT LQR} are satisfied.

\subsection{Double Integrator}
\begin{figure}
     \centering
     \begin{subfigure}[b]{0.22\textwidth}
         \includegraphics[width=\textwidth]{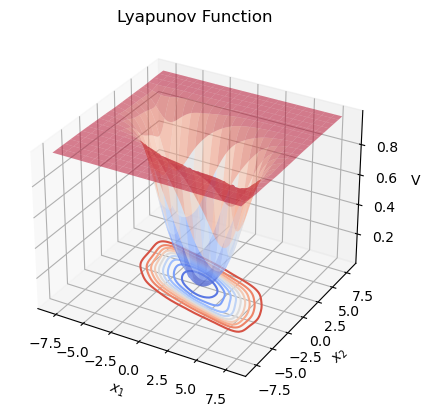}
         \caption{$W_{\theta}(x)$ learned for double integrator system}
         \label{fig:dint_3d}
     \end{subfigure}
     \begin{subfigure}[b]{0.22\textwidth}
         \includegraphics[width=\textwidth]{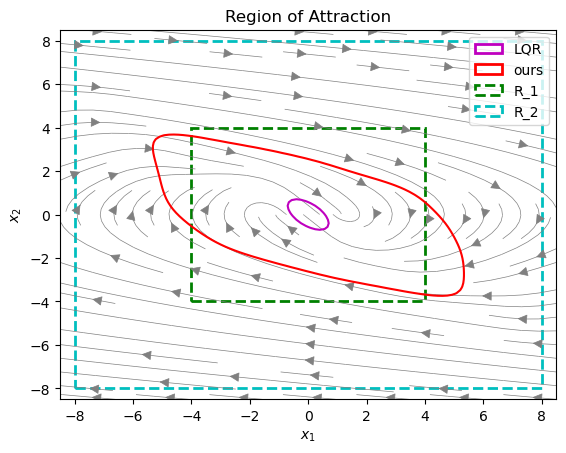}
         \caption{Verified DoA and vector field $\dot{x} = f(x, \pi_\gamma(x))$}
         \label{fig:dint_2d}
     \end{subfigure}
        \caption{Double integrator system}
        \label{fig:double integrator}
\end{figure}

We first consider the double integrator dynamics $\dot{x}_1 = x_2 ,\ \dot{x}_2=u$.
The learned Lyapunov function is shown in Figure \ref{fig:dint_3d}.
The verified DoA and the vector field induced by the learned controller are shown in Figure \ref{fig:dint_2d}. 

\subsection{Van der Pol}
\begin{figure}
     \centering
     \begin{subfigure}[b]{0.22\textwidth}
         \includegraphics[width=\textwidth]{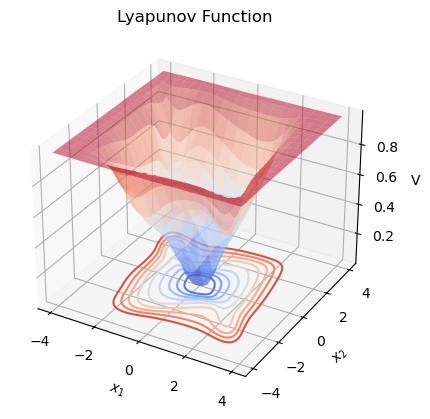}
         \caption{$W_{\theta}(x)$ learned for Van der Pol system}
         \label{fig:vanderpol_3d}
     \end{subfigure}
     \begin{subfigure}[b]{0.22\textwidth}
         \includegraphics[width=\textwidth]{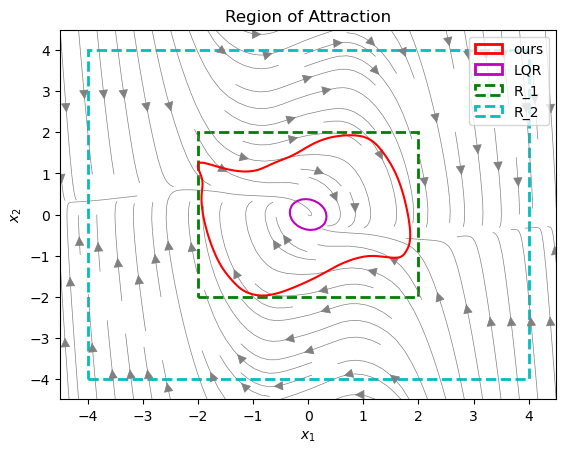}
         \caption{Verified DoA and vector field $\dot{x} = f(x, \pi_\gamma(x))$}
         \label{fig:vanderpol_2d}
     \end{subfigure}
        \caption{Van der Pol system}
        \label{fig:vanderpol}
\end{figure}

Consider the Van der Pol system  $\dot{x}_1 = x_2$ and $\dot{x}_2=x_1 - \mu (1 - x_1^2)x_2 + u$.
The learned Lyapunov function is shown in Figure \ref{fig:vanderpol_3d}.
The verified DoA and the vector field induced by the learned controller are shown in Figure \ref{fig:vanderpol_2d}.

\subsection{Inverted Pendulum}
\begin{figure}
     \centering
     \begin{subfigure}[b]{0.22\textwidth}
         \includegraphics[width=\textwidth]{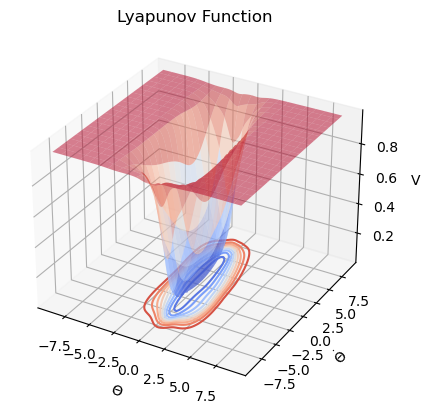}
         \caption{$W_{\theta}(x)$ learned for inverted pendulum system}
         \label{fig:pendulum_3d}
     \end{subfigure}
     \begin{subfigure}[b]{0.22\textwidth}
         \includegraphics[width=\textwidth]{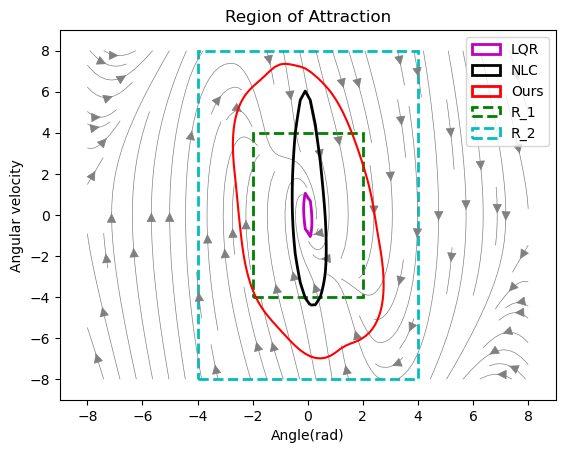}
         \caption{Verified DoA and vector field $\dot{x} = f(x, \pi_\gamma(x))$}
         \label{fig:pendulum_2d}
     \end{subfigure}
        \caption{Inverted Pendulum system}
        \label{fig:pendulum}
\end{figure}

The inverted pendulum system has two states, the angular position $\theta$, angular velocity $\omega$, and control input $u$. The dynamical system of inverted pendulum is $\dot{\theta} = \omega$ and $\dot{\omega}=\frac{g}{l} \sin(\theta) - \frac{b\omega}{ml^2} + \frac{u}{ml^2}$.
The learned Lyapunov function is shown in Figure \ref{fig:pendulum_3d}.
The verified DoA and the vector field induced by the learned controller are shown in Figure \ref{fig:pendulum_2d}.

\subsection{Bicycle Tracking}
\begin{figure}
     \centering
     \begin{subfigure}[b]{0.22\textwidth}
         \includegraphics[width=\textwidth]{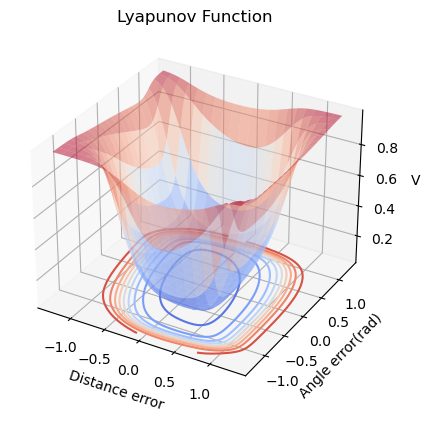}
         \caption{$W_{\theta}(x)$ learned for bicycle tracking system}
         \label{fig:tracking_3d}
     \end{subfigure}
     \begin{subfigure}[b]{0.22\textwidth}
         \includegraphics[width=\textwidth]{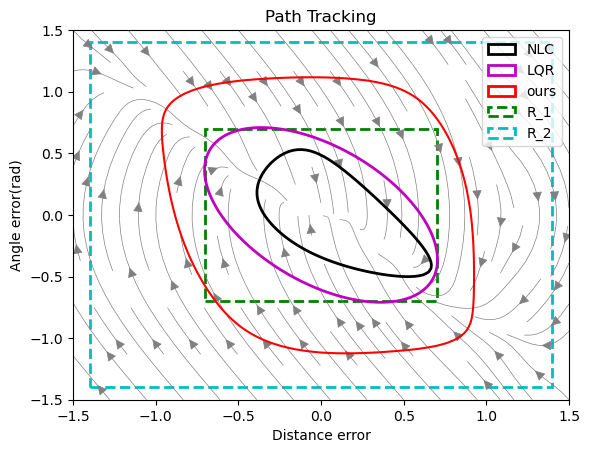}
         \caption{Verified DoA and vector field $\dot{x} = f(x, \pi_\gamma(x))$ }
         \label{fig:tracking_2d}
     \end{subfigure}
        \caption{Bicycle tracking system}
        \label{fig:tracking}
\end{figure}

The bicycle tracking system has two states, the distance $d_e$, angle error $\theta_e$, and control input $u$. The dynamical system for the tracking problem is $\dot{d}_e =  v \sin(\theta_e)$ and $\dot{\theta}_e = v  \frac{\tan(u)}{l} - \frac{\cos(\theta_e)}{1 - d_e}$. 
The learned Lyapunov function is shown in Figure \ref{fig:tracking_3d}.
The verified DoA and the vector field induced by the learned controller are shown in Figure \ref{fig:tracking_2d}.

In summary, all the experiments show that our verified DoA is 2$\sim$4 times larger than other methods. 
\begin{table}[]
\begin{tabular}{|l|l|l|l|l|}
\hline
 Dynamical System&  $W_\theta$ dimension &  $\pi_\gamma$ dimension & $\alpha$ & $c$ \\ \hline
 Double Integrator & [2,20,20,1] & [2, 10, 10, 1] &  0.05& 0.7  \\ 
 \hline
 Van der Pol & [2,30,30,1] & [2, 30, 30, 1]  & 0.1 & 0.5  \\ 
 \hline
 Inverted Pendulum & [2, 20, 20, 1] & [2, 5, 5, 1] & 0.2 & 0.7  \\
 \hline
 Bicycle Tracking & [2, 20, 20, 1] & [2, 10, 10, 1] & 1.5 & 0.4 \\
 \hline
\end{tabular}
\caption{Experiment Details}
\label{table:hyperparameters}
\end{table}

\section{CONCLUSIONS}
In this paper, we developed an actor-critic framework to jointly train a neural network control policy and the corresponding stability certificate, with the explicit goal of maximizing the induced region of attraction by leveraging Zubov's partial differential equation to inform the loss function of the shape of the boundary of the true domain of attraction. Our numerical experiments on several nonlinear benchmark examples corroborate the superiority of the proposed method over competitive approaches in enlarging the domain of attraction. For future work, we will incorporate robustness to model uncertainty in our loss function. We will also investigate the extension of the proposed method to discrete-time systems \cite{wu2024neural}.

\bibliographystyle{ieeetr}

\newpage
\appendix

\subsection{Lorenz System}
\begin{figure}[h]
     \centering
     \begin{subfigure}[b]{0.22\textwidth}
         \centering
         \includegraphics[width=\textwidth]{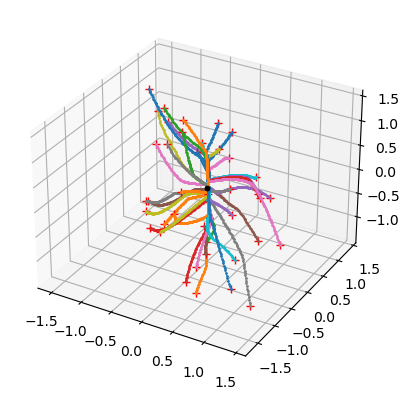}
         \caption{Sampled paths of Lorenz system with learned controller}
         \label{fig:lorenz_controlled}
     \end{subfigure}
     \hfill
     \begin{subfigure}[b]{0.22\textwidth}
         \centering
         \includegraphics[width=\textwidth]{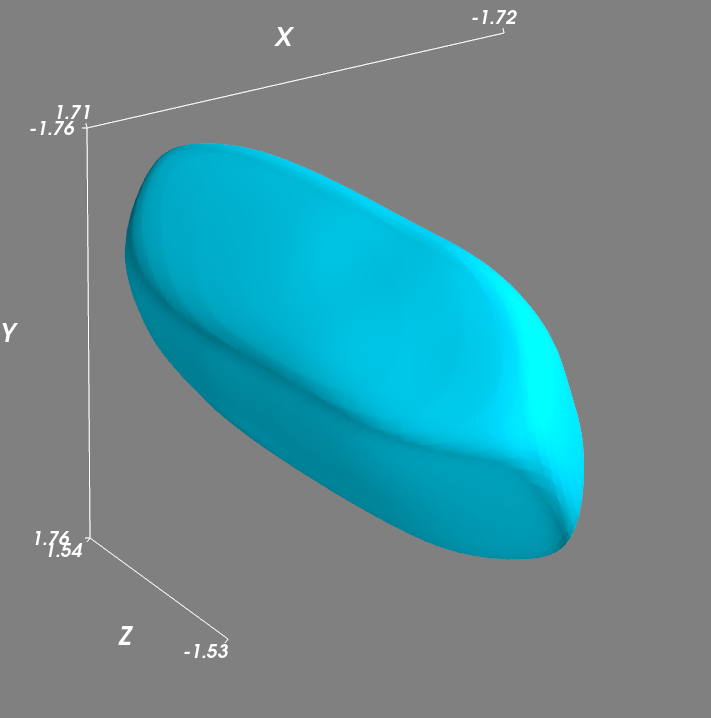}
         \caption{Verified DoA}
         \label{fig:lorenz_doa}
     \end{subfigure}
        \caption{3D Lorenz example }
        \label{fig:lorenz}
\end{figure}

We also provide a 3d example. Consider the Lorenz system with control input
\begin{align*}
    \dot{x_1} &= \sigma (x_2 - x_1) + u_1\\
    \dot{x_2} &= (r - x_3)x_1 - x_2 + u_2\\
    \dot{x_3} &= x_1 x_2 - bx_3 + u_3
\end{align*}
We use a typical choice $\sigma=10, r=28, b=\frac{8}{3}$. We consider control input $u_i \in [-10, 10] ~ \forall i=1,2,3$. Figure \ref{fig:lorenz_controlled} shows sampled trajectories with the learned controller. We managed to verify a $0.2$-level set of the learned Zubov function, and the level set is shown in Figure \ref{fig:lorenz_doa}. Since the verification is computationally expensive, we did not perform bisection to find the largest level set, thus the DoA shown in Figure \ref{fig:lorenz_doa} is an under-approximation of the largest verifiable DoA.

\subsection{Ablation Study}
\begin{figure}[h]
     \centering
     \begin{subfigure}[b]{0.15\textwidth}
         \centering
         \includegraphics[width=\textwidth]{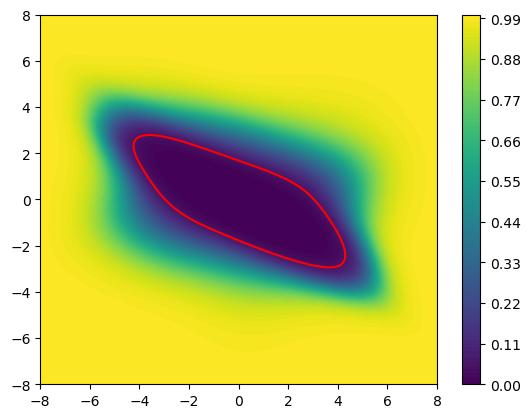}
         \caption{$\alpha=0.02$}
         \label{fig:al0.02}
     \end{subfigure}
     \hfill
     \begin{subfigure}[b]{0.15\textwidth}
         \centering
         \includegraphics[width=\textwidth]{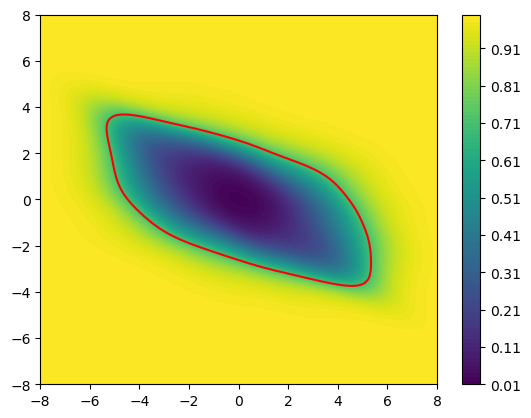}
         \caption{$\alpha=0.05$}
     \end{subfigure}
     \hfill
     \begin{subfigure}[b]{0.15\textwidth}
         \centering
         \includegraphics[width=\textwidth]{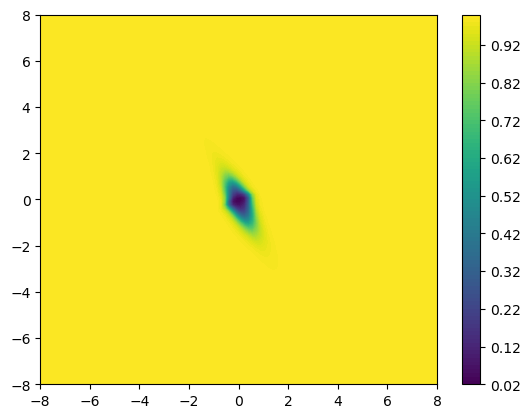}
         \caption{$\alpha=0.1$}
     \end{subfigure}
        \caption{Learned Zubov function and DoA for Double Integrator with different choice of $\alpha$}
        \label{fig:ablation_alpha}
\end{figure}

The choice of hyperparameter $\alpha$ is critical to the success of learning. Here we provide some intuition about the choice of $\alpha$.
We use the relationship $W(x) = \mathrm{tanh}(\alpha V(x))$ in 8(a), 
$\alpha$ is chosen so that $\alpha V(x)$ remains in the steep (non-flat) region of $\mathrm{tanh}$ for all $x \in R_1$. Thus, for large $R_1$, we choose smaller $\alpha$. 
For example, since in the double integrator example, $R_1$ is large and the trajectory tends to be long, we choose a small value $\alpha=0.05$. In the tracking example, $R_1$ is much smaller, so we choose a larger value $\alpha=1.5$. But in general, a small value (e.g. 0.05) works well across all examples. We have added an ablation for $\alpha$ in Figure \ref{fig:ablation_alpha}

\begin{figure}[h]
     \centering
     \begin{subfigure}[b]{0.15\textwidth}
         \centering
         \includegraphics[width=\textwidth]{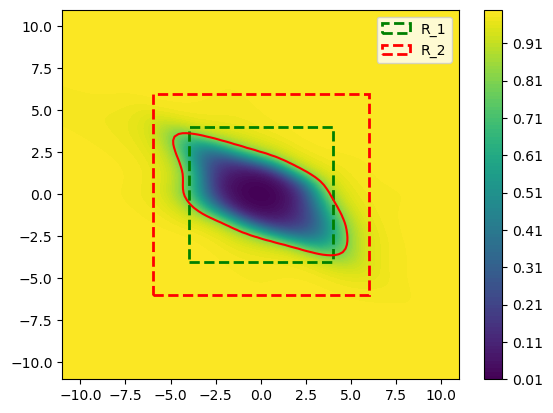}
         \caption{$a=1.5$}
     \end{subfigure}
     \hfill
     \begin{subfigure}[b]{0.15\textwidth}
         \centering
         \includegraphics[width=\textwidth]{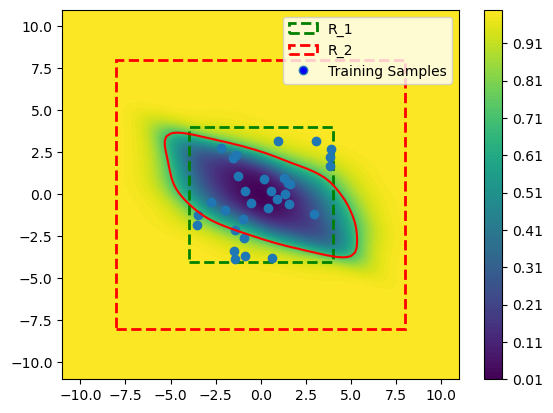}
         \caption{$a=2$}
     \end{subfigure}
     \hfill
     \begin{subfigure}[b]{0.15\textwidth}
         \centering
         \includegraphics[width=\textwidth]{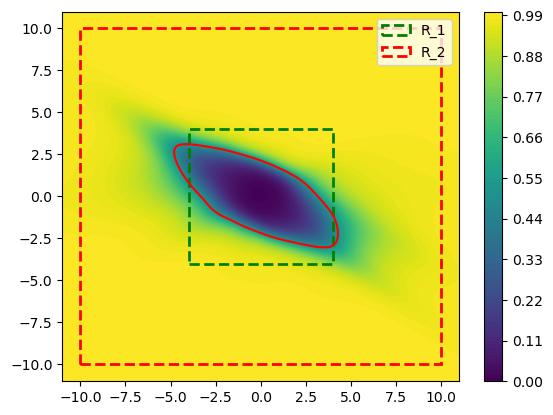}
         \caption{$a=2.5$}
     \end{subfigure}
        \caption{Learned Zubov function and DoA for Double Integrator with different choice of $a$ }
        \label{fig:ablation_a}
\end{figure}

Another hyperparameter is $a$, the ratio between $R_2$ and $R_1$. We choose $R_2 = 2*R_1$ in this paper. In principle, the choice of $a$ is flexible as long as $R_1$ is a subset of $R_2$. However, $R_1$ should be not much larger than the maximum possible DoA. Otherwise, most trajectories will not converge, and the training of $W(x) \approx \mathrm{tanh}(\alpha V)$ will have imbalanced samples so that $W(\cdot)$ collapses to 1. Given $R_2 = a*R_1$ for some $a>0$, we have done an ablation study on $a$ in Figure \ref{fig:ablation_a}. 

\end{document}